\documentclass[10pt,twocolumn,letterpaper]{article}

\font\cvprtenhv  = phvb at 8pt 
\font\elvbf  = ptmb scaled 1100

\usepackage{times}
\usepackage{epsfig}
\usepackage{graphicx}
\usepackage{amsmath}
\usepackage{amssymb}

\usepackage[pagebackref=true,breaklinks=true,letterpaper=true,colorlinks,bookmarks=false]{hyperref}

\usepackage{multicol}
\usepackage{lipsum}

\usepackage[version=4]{mhchem}
\usepackage{empheq}
\usepackage{mdframed}
\usepackage{nomencl,etoolbox,ragged2e,siunitx}

\newtheorem{prop}{Proposition}

\DeclareMathAlphabet\mathbfcal{OMS}{cmsy}{b}{n}

\usepackage{xcolor}
\title{Semi-Supervised Extensions}
\begin{document}

\title{Semi-Supervised Learning Enabled by\\Multiscale Deep Neural Network Inversion
}

\author{Randall Balestriero\\
ECE, Rice University USA\\
{\tt\small randallbalestriero@gmail.com}
\and
Herv\'e Glotin\\
LSIS, UTLN, France\\
\and
Richard G. Baraniuk\\
ECE, Rice UNiversity, USA\\
}
\date{}
\maketitle
\begin{abstract}
Deep Neural Networks (DNNs) provide state-of-the-art solutions in several difficult machine perceptual tasks. 
However, their performance relies on the availability of a large set of labeled training data, which limits the breadth of their applicability.
Hence, there is a need for new {\em semi-supervised learning} methods for DNNs that can leverage both (a small amount of) labeled and unlabeled training data.
In this paper, we develop a general loss function enabling DNNs of any topology to be trained in a semi-supervised manner without extra hyper-parameters.
As opposed to current semi-supervised techniques based on topology-specific or unstable approaches, ours is both robust and general.
We demonstrate that our approach reaches state-of-the-art performance on the SVHN ($9.82\%$ test error, with $500$ labels and wide Resnet) and CIFAR10 ($16.38\%$ test error, with $8000$ labels and sigmoid convolutional neural network) data sets.
\end{abstract}

\section{Introduction}

A {\em deep neural network} (DNNs) processes a signal $x$ to produce an inference by composing $L$ parametric mappings  $f^{(L)}_{\theta^{(\ell)}},\dots, f^{(1)}_{\theta^{(1)}}$ called layers, each with respective internal parameters $\theta^{(\ell)}$. 
Each layer $f^{(\ell)}_{\theta^{(\ell)}}$ takes an input volume $z^{(\ell-1)}(x)$ to create an output volume $z^{(\ell)}(x),\ell=1,\dots,L$ with $z^{(0)}(x)=x$
%
\begin{align}
z^{(\ell)}(x)= \left(f^{(\ell)}_{\theta^{(\ell)}} \circ \dots \circ f^{(1)}_{\theta^{(1)}}\right)(x), \quad \ell = 1,\dots,L.
\end{align}
Composing $L$ layers thus generates a collection of volumes until the final output $z^{(L)}(x)$ is reached.
%
%
%

For the case of {\em classification} with $C$ classes, on which we focus in this paper, the final DNN output $z^{(L)}(x)$ is transformed into a probability distribution via the softmax nonlinearity $S:\mathbb{R}^C\rightarrow \mathbb{R}^C$ to create the final prediction $\widehat{y}(x)=S(z^{(L)}(x))$, where
\begin{align}
\hat{y}(x)_c=\frac{e^{z^{(L)}(x)_c}}{\sum_{c=1}^Ce^{z^{(L)}(x)_c}},c=1,\dots,C.
\end{align}
The prediction corresponds to a class membership probability of $x$ belonging to class $c$.

Given a (large) set of input/output pairs 
$\mathcal{D}_s=\{(x_n,y_n)_{n=1}^{N_s}\}$ the parameters  $\Theta=\{\theta^{(1)},\dots\theta^{(L)}\}$ of the DNN are learned by comparing the prediction $\widehat{y}(x_n)$ to the target $y_n$ via a {\em loss function} such as cross-entropy $\mathcal{L}_{\rm CE}$ \cite{shore1981properties}. 
The induced error is then minimized by updating the parameters $\Theta$ via first-order techniques such as gradient descent \cite{larochelle2009exploring} leveraging backpropagation  \cite{hecht1988theory}. 

While powerful, the application of fully supervised learning framework is limited by the often high cost of obtaining the required (very) large labeled labeled training dataset.
Consequently, there is growing interest in leveraging unlabeled data $\mathcal{D}_{\rm u}=\{(x_n,\emptyset)_{n=1}^{N_u}\}$, which are often abundant.   
The focus of this paper is on fusing the information present in both $\mathcal{D}_s$ and $\mathcal{D}_{\rm u}$ to effect {\em semi-supervised learning} with DNNs.

Deriving an semi-supervised learning framework that is robust and most importantly architecture agnostic allowing the use of resnet, and with few hyper-parameters remains an open problem in deep learning.

In this paper, we tackle this challenge by introducing a new semi-supervised learning framework for DNNs.
The framework proposes first a renormalization of the original semi-supervised loss presented in \cite{dnninversion} as well as a multiscale reconstruction loss that contributes stability during learning by reducing the impact of noisy or corrupted inputs. The term inversion is used loosely as the inversion problem in a nonlinear DNN is an ill posed problem in general. Thus in our case $f^{(-1)}$ is used in place of the ill-defined inverse.
We summarize our {\bf major contributions} as:
\begin{itemize}
\setlength{\itemsep}{0mm}
    \item A thorough analysis of the loss function presented in \cite{dnninversion} and {\em removal of all hyper-parameters} via a loss-dependent renormalization that obviates fine hyper-parameter cross-validation (see Section \ref{sec:renorm}).
    
    \item Introduction of a new {\em \textbf{multiscale} loss for semi-supervised learning} that is robust to initialization, the sampling of the labeled dataset $\mathcal{D}$, and the presence of noise in the input (see Section \ref{sec:layer}).
    
    \item A series of {\em exhaustive experiments} with the SVHN and CIFAR10 datasets and multiple DNN topologies that demonstrate that our approach achieves state-of-the-art results (see Section \ref{sec:exp}).
    
\end{itemize}


\textbf{Related Work:}
The problem of semi-supervised learning with DNN has been attempted by several groups.
The improved {\em generative adversarial network} (GAN) technique \cite{salimans2016improved} couples two deep networks: a generative model creating new signal samples, and a discriminative model performing supervised learning. The discriminator simultaneously performs two tasks: discriminating between the true sample distribution and the generated one, and classifying the labeled samples from $\mathcal{D}_s$.
{\em Triple Generative Adversarial Nets} \cite{li2017triple} propose a extension of the GAN framework for the particular task of semi-supervised by introduction of a third player. The task thus becomes simpler as there exist one discriminator labeling images (fake or real) and another predicting if the couples (image,label) are fake or not. Through this, better stability is reached. Finally, {\em Good Semi-supervised Learning That Requires a Bad GAN} \cite{dai2017good} currently hold SOTA method. This work lessen the same problem of GAN for semi-supervised of \cite{li2017triple} by deriving analytical conditions and better formulation for the GAN objective hence providing a finer loss function as opposed to a third network.

The {\em probabilistic formulation of deep convolutional networks} presented in \cite{patel2016probabilistic} supports semi-supervised learning. However, due to the need to have tractable probabilistic graphical model (PGMs), many simplifications led to this approach being applicable exclusively with Deep Convolutional Networks (DCN) topologies with Relu and max-pooling. Also, it requires the inputs $x$ and inner representations $z^{(\ell)},\ell=1,\dots,L $ to be non-negative, making most general tasks out of reach.
Temporal Ensembling for Semi-Supervised Learning \cite{laine2016temporal} propose to constrain the representations of a same input stimuli to be identical in the latent space despite the presence of dropout noise. This search of stability in the representation is analogous to the one of a siamese network \cite{hoffer2015deep} but instead of presenting two different inputs, the same is used through two different models (induced by dropout). This technique provides an explicit loss for the unsupervised examples leading to the $\Pi$ model just described and a more efficient method denoted as temporal ensembling.
Distributional Smoothing with Virtual Adversarial Training \cite{miyato2015distributional} proposes also a regularization term constraining the regularity of the DNN mapping for a given sample. Based on this a semi-supervised setting is derived by imposing for the unlabeled samples to maintain a stable DNN. Those two last described methods are the closest one of the proposed approach in this paper for which, the DNN stability will be replaced by a reconstruction ability, closely related to the DNN stability.

Classical approaches when considering the options for DNN inversion  was provided in \cite{dua2000inversion} and in general relate to flavors of autoencoders \cite{ng2011sparse}, such as the stacked convolutional autoencoder \cite{masci2011stacked}.
As such, the {\em semi-supervised with ladder network} approach \cite{rasmus2015semi} can be seen as a particular autoencoder. It employs a \textbf{per-layer} reconstruction loss defined as $\mathcal{L}_{R}^{(\ell)}(x) = ||z^{(\ell)}(x)- \frac{ d z^{(\ell+1)}(x)}{d z^{(\ell)}(x)}^T z^{(\ell+1)}(x) ||^2,\ell = L-1,\dots, 0$. In the latter equation, $\frac{ d z^{(\ell+1)}(x)}{d z^{(\ell)}(x)}$ represents the derivative of the representation of the $\ell+1$ layer w.r.t. the previous layer representation.
By forcing the inner layer to output an encoding describing the class distribution of the input via softmax nonlinearity, this deep unsupervised model is turned into a semi-supervised model. There remains a lack of a path to generalize this approach to other network topologies, such as recurrent or residual networks. Also, the per-layer ''greedy'' reconstruction loss might be sub-optimal unless correctly weighted pushing the need for a precise and large cross-validation of hyper-parameters.

Other attempts based on back-propagation such as in \cite{zeiler2010deconvolutional,zeiler2014visualizing} provides working solutions and efficient implementations, yet, did not leverage the approach for semi-supervised learning. In addition, generalization those layer specific technique to any architecture is not clear.
 However, recent work on {\em DNN inversion} \cite{dnninversion} has developed a general approach applicable to any topology. In particular, they presented  semi-supervised state-of-the-art results on MNIST via the use of a Resnet topology. To do so, they introduce a generic way to invert a given DNN and  define a \textbf{global} reconstruction loss $\mathcal{L}_{R}^{(\ell)}(x) = ||x- \frac{ d z^{(L)}(x)}{dx}^Tz^{(L)}(x) ||^2$, as well as an entropy loss $\mathcal{L}_E(\hat{y}(x))=-\sum_{c=1}^C\hat{y}(x)_c \log(\hat{y}(x)_c)$ for the unlabeled examples. One can notice the difference from the ladder network by defining a global reconstruction loss as opposed to per-layer.

For all the presented method, two main drawbacks arise. The first one is the presence of hyper-parameters to combine the different losses. The second, comes from the reconstruction loss. In the presence of noise, or corrupted inputs, the reconstruction objective will lead to noisy weights updates for all parameters $\theta^{(\ell)},\ell=1,\dots,L$ slowing convergence and hurting final performances.

\section{A Universal \& Robust Semi-Supervised Loss}

In order to overcome the input sensitivity of the reconstruction loss as well as the need for fine cross-validation by hyper-parameter removal. 
We first introduce notations and review the original scheme of \cite{report,dnninversion}.
Afterwards, we will develop a simple loss-dependent renormalization that makes the loss's behavior invariant to the task and topology at hand. 
We will robustify our method by modifying the reconstruction loss leading to greater stability for real world datasets (as we demonstrate below in Sec.~\ref{sec:exp}).

\subsection{Multi-Objective Loss Renormalization}\label{sec:renorm}

The work on semi-supervised learning for DNNs proposed in \cite{dnninversion} leverages the inverse DNN formula defined as $f^{-1}(x):=\frac{df(x)}{dx}^Tf(x)$. Based on this, a reconstruction loss has been defined as 
\begin{align}
\mathcal{L}_{R}(x) = \left\|x- \frac{ d z^{(L)}(x)}{dx}^Tz^{(L)}(x) \right\|^2.
\end{align}
This loss acts as a data-driven network regularizer such that information of unlabeled samples is taken into account in the way DNNs model their input \cite{report}. This is opposed to the standard structural regularization such as Tikhonov penalty \cite{tikhonov1966stability}. Additionally, an entropy loss was defined for the unlabeled samples as 
\begin{align}
\mathcal{L}_E(\hat{y}(x))=-\sum_{c=1}^C\hat{y}(x)_c \log(\hat{y}(x)_c).
\end{align}
The presence of the entropy loss applied on the unlabeled data is natural. In fact, for supervised labels, the optimal output distribution is the one of minimum entropy (i.e., Dirac) constrained such that the position of this energy impulse is at the right index (class) position. For unsupervised examples, while this index position is unknown, the optimal remains a distribution of minimal entropy. Hence, $\mathcal{L}_E$ acts as a guide, or attention model, on the internal parameters pushing unsupervised examples towards a known labeled cluster learned via the cross-entropy $\mathcal{L}_{CE}$ and $\mathcal{D}_s$. As a result, the final semi-supervised loss is a convex combination of the three losses defined as 
\begin{align}\label{original_eq}
\mathcal{L}(x,y)=&\alpha 1_{\{y\not = \emptyset \}} \mathcal{L}_{CE}(x,y)+(1-\alpha)\beta1_{\{y = \emptyset \}}\mathcal{L}_E(x)\nonumber \\
&+(1-\alpha)(1-\beta) \mathcal{L}_{R}(x).
\end{align}
The coefficients $\alpha,\beta \in [0,1]^2$ represent the weighting of the supervised versus unsupervised losses as well as regularization versus clustering. However, cross-validation of those parameters is cumbersome and heavy on computational power.
This, we renormalize each of the losses to ensure that their impact is equally distributed w.r.t. the overall loss. The following renormalized {\em global} loss function is proposed
\begin{align}\label{eq_renorm}
\mathcal{L}(x,y&)=\frac{1}{\log(C)}\Big(\underbrace{1_{\{y\not = \emptyset \}} \mathcal{L}_{CE}(x,y)}_{\text{Supervised Cluster Labeling}}\nonumber \\
&+\underbrace{1_{\{y = \emptyset \}}\mathcal{L}_E(x)}_{\text{Unsupervised Clustering}}\Big)+\frac{1}{D}\underbrace{ \mathcal{L}_{R}(x)}_{\text{Input Reconstruction}},
\end{align}
with $D$ the dimensionality of the input $x$.
The two losses $\mathcal{L}_E$ and $\mathcal{L}_{CE}$ are of same amplitude order. In both cases, we have at initialization $\mathcal{L}_{CE}(x,y)\approx \log(C)$ and $\mathcal{L}_{E}(x)\approx \log(C)$ as $\hat{y}(x) \sim \mathcal{\pi}(C)$ with $\mathcal{\pi}(C)$ Dirichlet distribution with uniform parameters. For the reconstruction loss $\mathcal{L}_{R}$, the range depends on the infinite norm of the considered input $x$. As we set in the experiments $||x||_\infty=1$, we ensure that this loss lies in the same range of values as the cross-entropy and entropy one. Also, due to the standard weight initialization of the layers, at initialization, reconstruction should not reach high amplitude values per pixel. This makes the three losses behaving with the same regime. We now propose further extension of this loss and specifically the reconstruction loss to provide stable and robust performances when dealing with real world datasets.

\begin{figure}[!tb]
    \centering
  \includegraphics[width=0.99\linewidth]{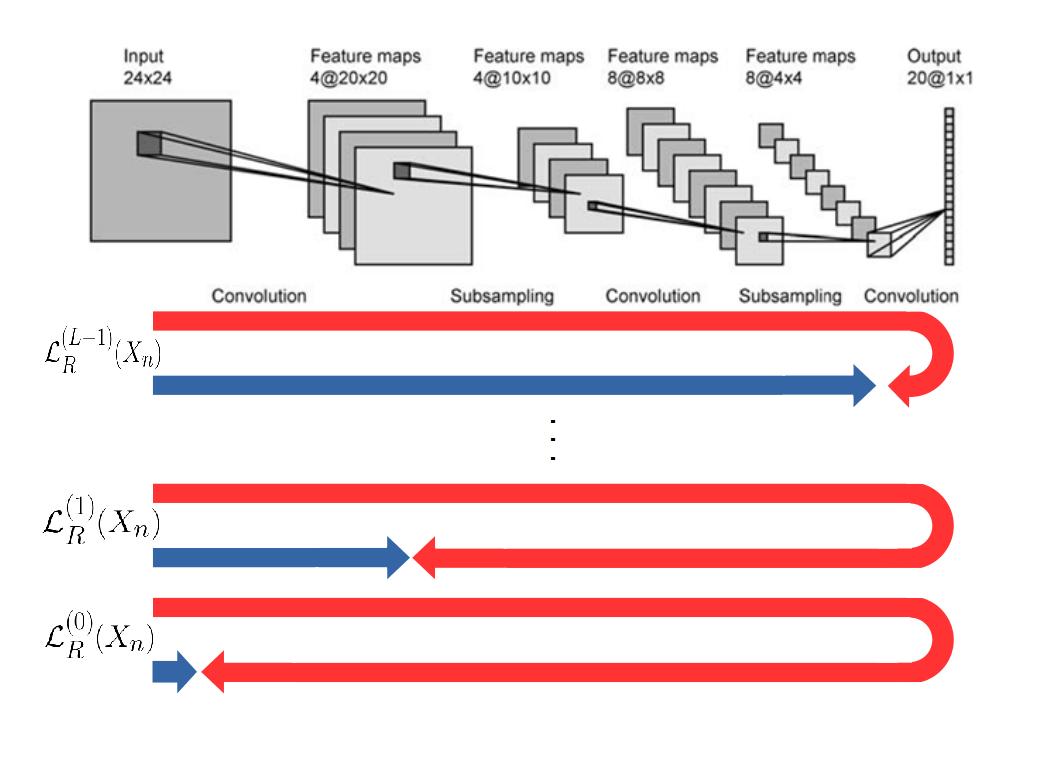}
  \vspace*{-3mm}
    \caption{The blue path corresponds to the forward inference computation through the DNN, while the red path represents the backward reconstruction. For each layer, those two representations are compared via $\left\|\color{blue} z^{(\ell)}\color{black} -\color{red} \frac{ d z^{(L)}}{d z^{(\ell)}}^T z^{(L)}\color{black} \right\|^2$.  There is no additional computational complexity in computing this per-layer error, since it uses the same operations as standard backpropagation.}\label{fig:per_layer_loss} 
\end{figure}

\subsection{Robust Semi-Supervised Learning via Multiscale  Reconstruction Loss}\label{sec:layer}

By introducing a finer reconstruction loss we aim to make performances robust to input noise and corruption as well as providing stable gradient updates. To do so we replace the global reconstruction loss based only on the input and its reconstruction by a convex combination of all the inner layers reconstructions. We do so for all the inner representations $z^{(\ell)},\ell=0,\dots,L-1$ including the input and excluding the final output. Let first define the per layer reconstruction loss as
\begin{align}
\mathcal{L}_{R}^{(\ell)}(x) =& \left\|z^{(\ell)}(x)- \frac{ d z^{(L)}(x)}{dz^{(\ell)}(x)}^Tz^{(L)}(x) \right\|^2,
\end{align}
for $\ell = L-1,\dots, 0$. In order to provide renormalization of each of those local losses we first remind briefly standard notations. Each of the generated DNN volumes $z^{(\ell)},\ell=1,\dots,L$ is of shape  $(C^{(\ell)},I^{(\ell)},J^{(\ell)})$. We denote by $D^{(\ell)}$ the total size of the $\ell^{th}$ volume defined as $D^{(\ell)}=C^{(\ell)}I^{(\ell)}J^{(\ell)}$.
Hence the {\em local} loss is defined as $\mathcal{L}(x,y)$ by replacing the normalized reconstruction term $\frac{1}{D}\mathcal{L}_R(x)$ with introduced re-normalized per layer reconstruction as
\begin{align}
\underbrace{\frac{1}{D}\mathcal{L}_R(x)}_{\text{Global}} \rightarrow \underbrace{\frac{1}{L}\sum_{\ell=0}^{L-1} \frac{1}{D^{(\ell)}}\mathcal{L}^{(\ell)}_{R}(x)}_{\text{Local/Hierarchical}}
\end{align}
Doing so, we have the following property making this reconstruction loss robust and stable for general tasks. For clarity we will now denote by $\Gamma$ and  $\lambda$ the global and local reconstruction losses as
\begin{align}\label{eq_phis}
    \Gamma(x) =& \frac{1}{D}\mathcal{L}_R(x)\\
    \lambda(x) =& \frac{1}{L}\sum_{\ell=0}^{L-1} \frac{1}{D^{(\ell)}}\mathcal{L}^{(\ell)}_{R}(x)
\end{align}

\begin{prop}
Given the local reconstruction loss $\lambda$, the impact of corrupted or noisy inputs is inversely proportional to the number of layers in the DNN.
\end{prop}
This result is direct since we have $\frac{1}{L}\mathcal{L}_R^{(0)}(x) \rightarrow 0$ as $L$ increases. 
We present in Fig.~\ref{fig:per_layer_loss} a depiction of the process with the blue arrow representing the forward pass, the red the reconstruction and this for all the layers including the input considered as layer $0$.

Hence, the impact of incorrect input normalization or presence of noise will only induce noisy gradients for the updates of $\theta^{(\ell)},\ell=1,\dots,L$ via the erroneous term $\frac{1}{L}\mathcal{L}_R^{(0)}(x)$. This induced noisy gradient will then be overcome by the induced ones from the inner layers reconstruction loss. Doing so, a DNN will maintain inner layer stability even if this implies an incorrect input reconstruction. We observe this exact behavior and the explosion of inner layer regularity in the experiment section where we provide and analysis the evolution of the losses for the $\lambda$, and $\Gamma$ settings.
The way we defined the per layer loss might seem arbitrary as opposed to the other possibility being $\mathcal{L}^{(\ell)}_R(x)=|| x-\frac{dz^{(\ell)}}{dx}z^{(\ell)} ||,\ell=1,\dots,L$. In this latter case, there is a per layer loss. Yet, in the presence of noise, corruption or simply class independent information in $x$,  this loss will provide noisy updates to all inner layers with the same impact disregarding of the number of layers. Hence, as it is the case for most application, with for example background, measurement noise and so on, our proposition is the one that should be chosen to ensure that these perturbations do not impact negatively the learning. Yet information of unlabeled examples are taken into account. Heuristically, it is observed that inner representation, by being the result of succession of mappings and nonlinearities will contain less and less class independent information. Hence, pushing reconstruction of inner representations as opposed to the input should be considered as the optimal strategy for real world application.

\section{Experimental Results}\label{sec:exp}

We first emphasize the need for cross validation encountered in the original framework. In order to reach state-of-the-art results on MNIST with $50$ labels, as we report the results in Tab.~\ref{mnist}, one should note the selected hyper-parameters $(\alpha,\beta)$. The best result was obtained by reducing the importance of the unsupervised losses likely due to the impacts of the reconstruction loss $\mathcal{L}_R$ becoming detrimental for the behavior of the whole DNN training.

\begin{table}
\caption{MNIST Dataset experiment demonstrating the importance of the loss weighting and the need to reduce the impact of $\mathcal{L}_R$ as $\alpha>0.5$ and $\beta<0.5$ in Eq.~\ref{original_eq}.}
\centering\label{mnist}
\begin{tabular}{|c|c|} \hline
$\mathbf{N_L} $             & $\mathbf{50}$  \\ \hline
Resnet2-32max with $\Gamma$ \cite{dnninversion}          &  $\textbf{99.14}$\\ &$(\alpha=0.7,\beta=0.2)$    \\ 
\hline 
Improved GAN \cite{salimans2016improved}        & $97.79 \pm 1.36$ \\ \hline
\end{tabular}
\end{table}

The optimal parameters being in favor of the supervised loss and further reducing the impact of $\mathcal{L}_R$ by setting $\beta=0.2$ is indication of the need to have better behaving reconstruction loss.
This further motivates the need to adapt the loss in order to remove the need for semi-supervised specific cross-validation as well as prevent the unsupervised loss to overcome the natural learning of the DNN with the given labels. We now run experiments on the present framework with the $\lambda$ versus $\Gamma$ losses.

\subsection{Per Layer Reconstruction}
\begin{figure*}[!tbp]
  \centering
  \begin{minipage}[b]{0.24\textwidth}
  a)
    \includegraphics[width=\textwidth]{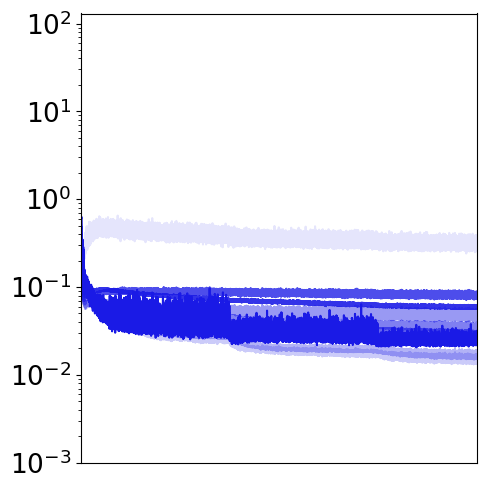}
  \end{minipage}
  \hfill
  \begin{minipage}[b]{0.24\textwidth}
    \includegraphics[width=\textwidth]{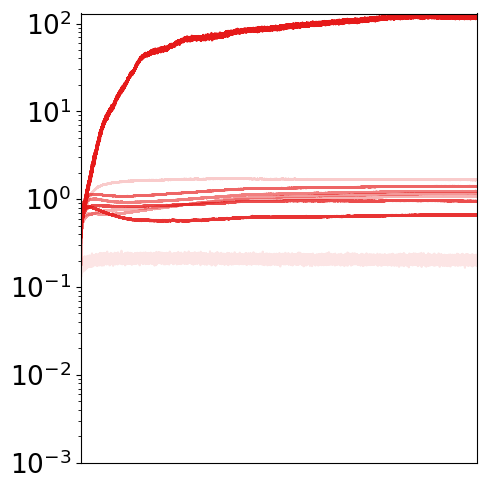}
   \end{minipage}
   \hspace{0.1cm}
  \begin{minipage}[b]{0.24\textwidth}
  b)
    \includegraphics[width=\textwidth]{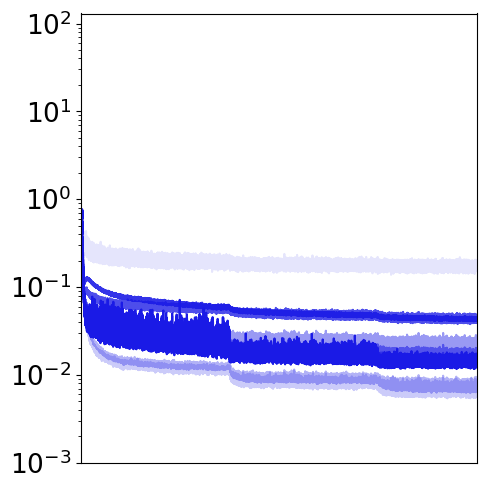}
  \end{minipage}
  \hfill
  \begin{minipage}[b]{0.24\textwidth}
    \includegraphics[width=\textwidth]{figs/CIFAR8000SEMI2LargeCNNmean-global_onlyloss.png}
   \end{minipage}
   \\ 
  \begin{minipage}[b]{0.24\textwidth}
  c)
    \includegraphics[width=\textwidth]{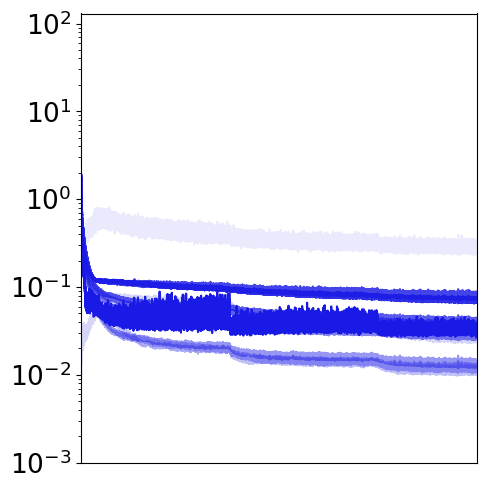}
  \end{minipage}
  \hfill
  \begin{minipage}[b]{0.24\linewidth}
    \includegraphics[width=\textwidth]{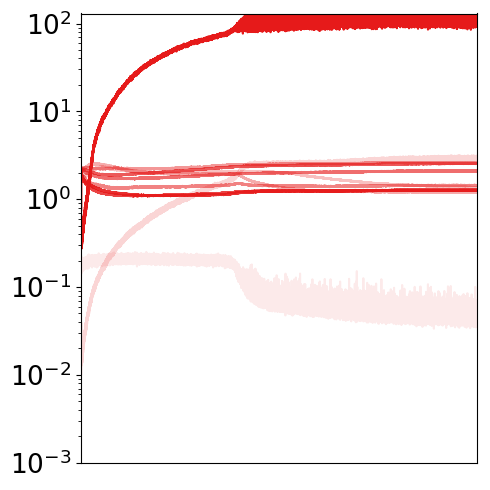}
   \end{minipage}
   \hspace{0.1cm}
     \begin{minipage}[b]{0.24\textwidth}
     d)
    \includegraphics[width=\textwidth]{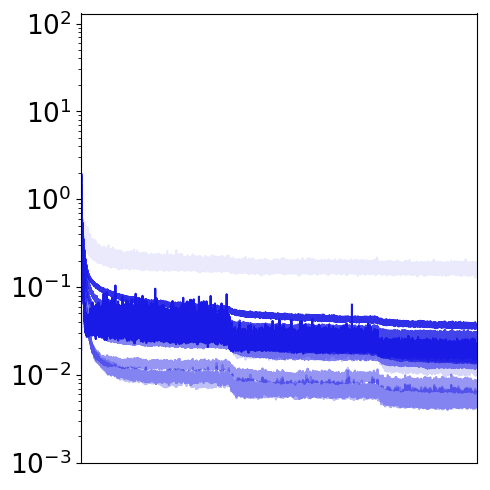}
  \end{minipage}
  \hfill
  \begin{minipage}[b]{0.24\textwidth}
    \includegraphics[width=\textwidth]{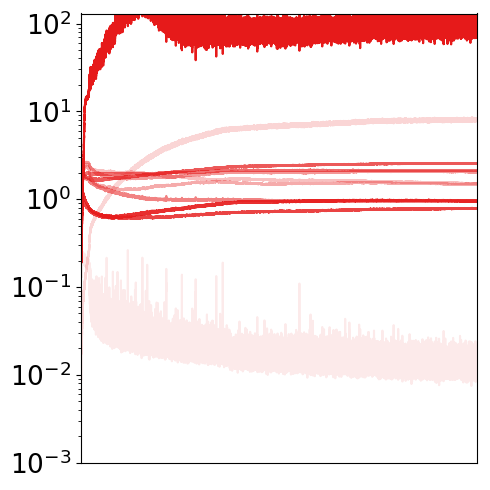}
   \end{minipage}
  
    \caption{a,c) CIFAR10 with 8000 labels task, b,d) SVHN with 1000 labels task. Top rows a,b) represent the CNN model with LReLU activation functions and bottom rows c,d) the wide resnet architecture. In blue is depicted the $\lambda$ loss and in red the $\Gamma$ loss during learning with darker colors for inner layers. The axis are aligned to provide better comparison ranging from $0.001$ to $130$. Clearly, the $\Gamma$ loss favorized the learning of internal representations with greater and greater mismatch between the forward-backward flow. On the other hand, the $\lambda$ loss provides stable representation at each inner layer. The difference in input reconstruction error $\mathcal{L}_R^{(0)}$ does not differ greatly due to the input renormalization. However, inner layer losses $\mathcal{L}_R^{(L-1)},\dots$ sees its error explodes for the $\Gamma$ loss. Concerning inter architecture analysis, we can clearly see the ability of the resnet to reduce its reconstruction error whether in the $\lambda$ or $\Gamma$ setting thanks to its linear connections.}\label{all}
\end{figure*}

\begin{figure*}[!tb]
  \centering
    \includegraphics[width=\textwidth]{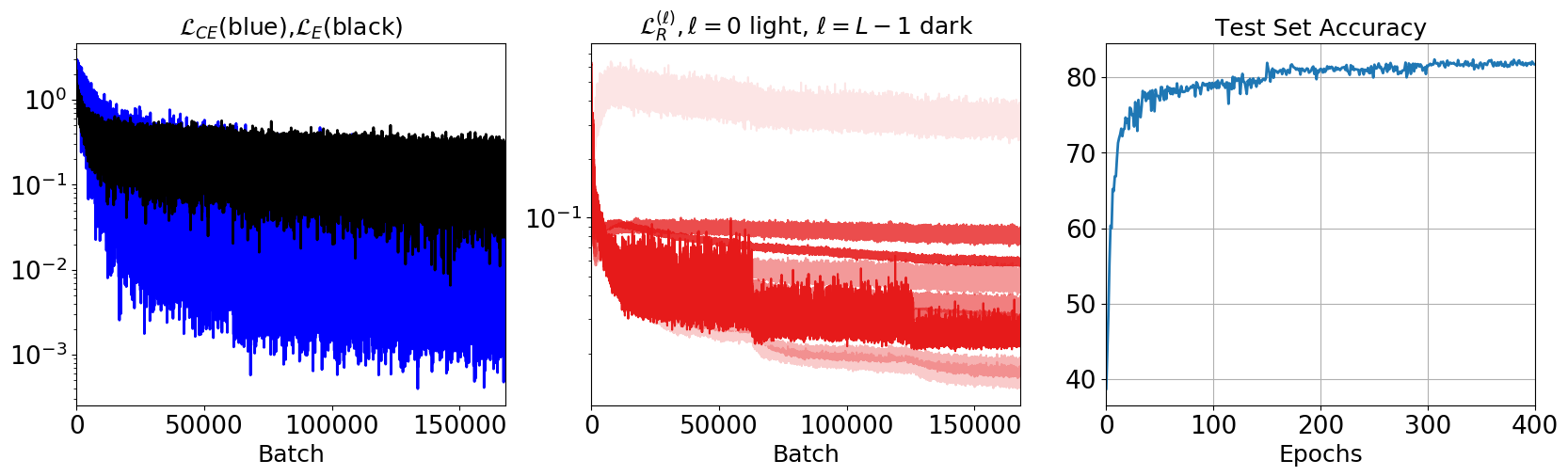}
    \caption{
    CIFAR10 with $8000$ labels,  $\lambda$ (multiscale) loss, CNN model with LReLU activation functions. Left: Cross-entropy (blue) and entropy (black) losses during learning for each batch in semilogy. Middle: Per layer losses evolution during training per batch with dark colors for inner layers in semilogy. Right: test set accuracy after each epoch computed as the categorical accuracy.}\label{losscifar}
\end{figure*}

\begin{figure*}[!tbp]
  \centering
    \includegraphics[width=\textwidth]{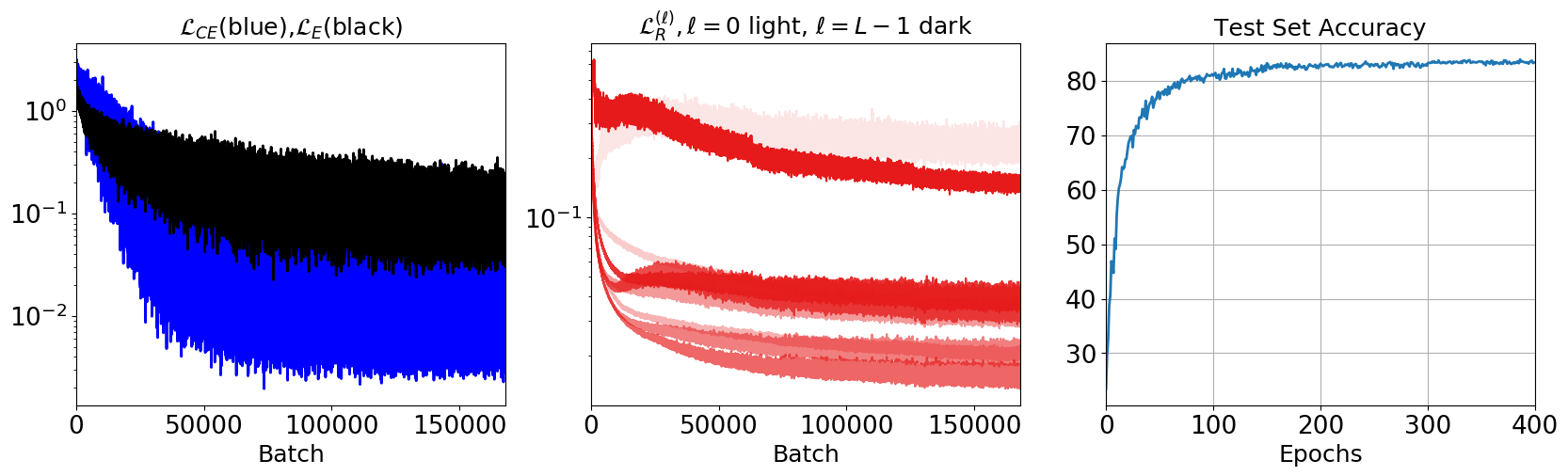}
    \caption{CIFAR10 with $8000$ labels,  $\lambda$ (multiscale) loss, CNN model with sigmoid activation functions. Left (semilog): Cross-entropy (blue) and entropy (black) losses during learning for each batch. Middle (semilog): Per layer losses evolution during training per batch with dark colors for inner layers. Right: test set accuracy after each epoch computed as the categorical accuracy.}\label{losscifarsigmoid}
\end{figure*}

\begin{figure*}[!tbp]
  \centering
  \begin{minipage}[b]{0.49\textwidth}
    \centering
  a) \\
    \includegraphics[width=\textwidth]{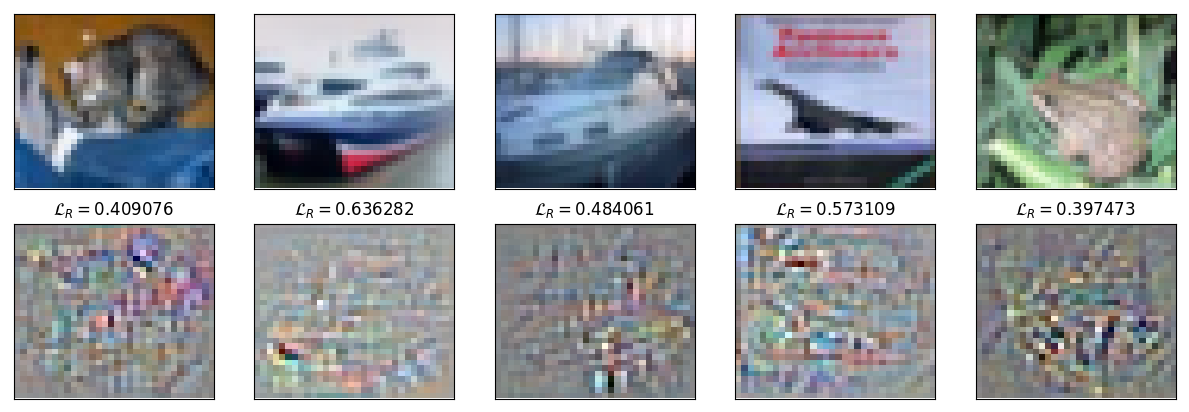}
  \end{minipage}
  \hfill
  \begin{minipage}[b]{0.49\textwidth}
     \centering
  b) \\
  \includegraphics[width=\textwidth]{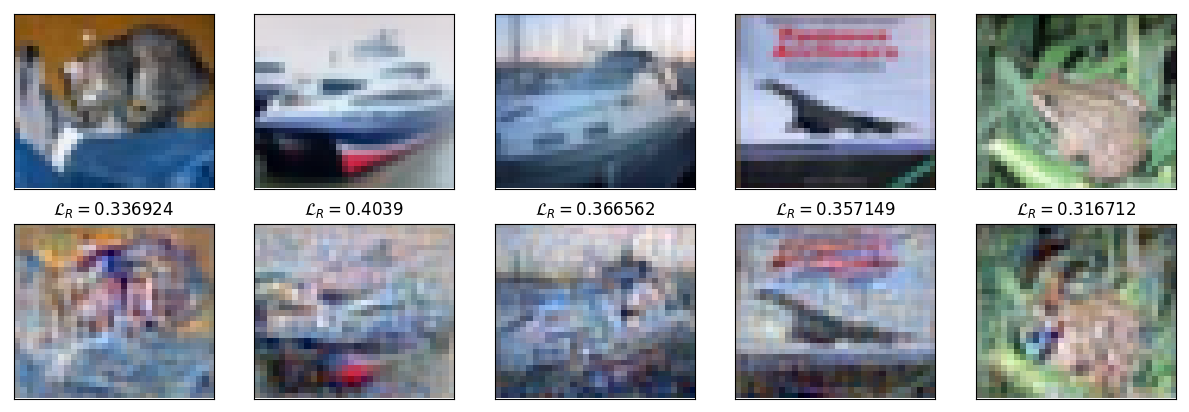}
   \end{minipage}
   \\ \vspace{2mm}
   
  \begin{minipage}[b]{0.49\textwidth}
     \centering
   c)\\
  \includegraphics[width=\textwidth]{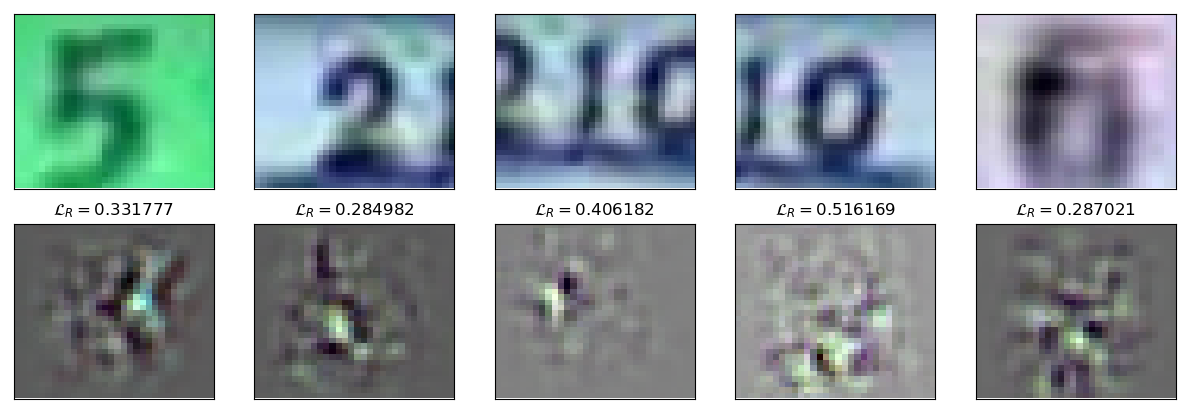}
  \end{minipage}
  \hfill
  \begin{minipage}[b]{0.49\textwidth}
     \centering
  d) \\
  \includegraphics[width=\textwidth]{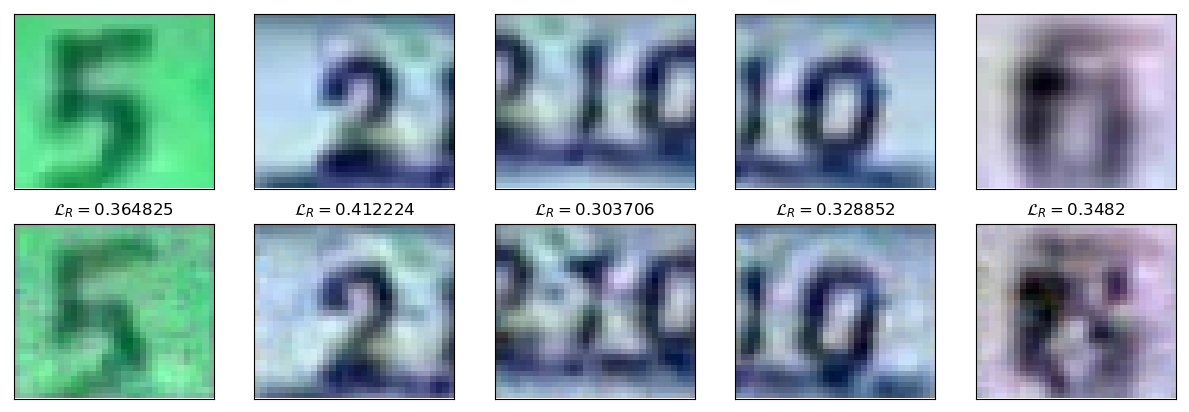}
   \end{minipage}
    \caption{Image reconstructions with the wide Resnet model when using the $\lambda$ (multiscale) loss for a,c) versus $\Gamma$ (global) loss for b,d) after training on CIFAR10 with $8000$ labels (top row) and SVHN with $1000$ labels (down row). For each, original image and reconstruction are presented with the reconstruction error $\mathcal{L}_R^{(0)}(x)$ in the titles. Clear distinctions can be noticed in the reconstruction ability of the network when changing from the multiscale to global loss. While the global loss is able to provide accurate reconstruction making humans able to identify the original label of the input, we demonstrate in Tab.~\ref{cifar},\ref{svhn} that this implies less classification capacities. In fact, when considering the plane image with red text on the top, it is clear that forcing a DNN to reconstruct this text will imply learning weights in a way that can not help for classification generalization. Hence, for real world images with noisy patterns such as background and noise, input reconstruction is detrimental for the classification task.}\label{rec}
\end{figure*}

In this section, we first describe the settings in which our experiments were performed. The Tab.~\ref{svhn},\ref{cifar} provide series of experiments on the two datasets SVHN and CIFAR, each time with two regime of labeled samples according to standard literature. We also provide evolution of the losses during training in Fig.~\ref{losscifar} and \ref{losscifarsigmoid}. Finally, image reconstruction is provided in order to qualitatively judge the abilities of the trained models to indeed reconstruct their input and provide further analysis between the $\lambda$ versus $\Gamma$ losses in Fig.~\ref{rec}.

To highlight the inter-dataset capacities of the model, we perform all experiments with different topologies but keeping identical the learning rate, batch size and input renormalization. 
We test $3$ different network topologies: a large CNN made of $9$ convolutional layers containing a total of  1M3 parameters; a wide Resnet with $2M1$ parameters denoted as Resnet3-64 ; a deep resnet with $1M1$ parameters denoted as Resnet6-32 with $M$ stading for million.
For the CNN, we provide for each layer the tuple (number of filters, shape of filters, padding, pooling size), with a pooling size of $1$ being synonym of no pooling performed. This leads, from the input layer to inner layer: $(96,3,s,1)$, $(96,3,f,1)$, $(96,3,f,2)$, $(192,3,v,1)$, $(192,3,f,1)$, $(192,3,v,2)$, $(192,3,v,1)$, $(192,1,s,6)$ where $s$ stands for same, $v$ for valid and $f$ for full. Finally, a fully connected layer with $10$ output neurons is used for the output prediction.
Note that this is a standard topology already used in \cite{patel2016probabilistic} for semi-supervised learning.
For the Resnet blocks, we use a simplified version of \cite{zagoruyko2016wide}. 
The Resnet block is defined as $f^{(\ell)}(z^{(\ell-1)})=Wz^{(\ell-1)}+f_{\rm conv}(z^{(\ell-1)})$. The operator $W$ is a linear convolution with filters of size $(1,1)$. We follow standard procedure as in \cite{zagoruyko2016wide} for the number of filters which are always of spatial size $(3,3)$ for the nonlinear convolutional layer and $(1,1)$ for the linear one. The number of filters is multiplied by $2$ after $n$ blocks, and at the same time a down-sampling of the representation by a factor of $(2,2)$ via mean pooling is performed. The total number of blocks is thus $3n$. The initial number of filters is denoted by $k$, then a full topology is written as Resnet$3$-$64$ for $n=3$ and $k=32$. Note that we used mean-pooling in the convolutional layer to prevent artifact due to the max-pooling when performing reconstruction.

In all cases, dropout \cite{srivastava2014dropout} is used after each nonlinearity with $p=0.2$ and batch norm \cite{ioffe2015batch} prior to nonlinearity taken as leaky-rectify \cite{xu2015empirical}.
All inputs $x_n$ are renormalized per observation by centering and reducing leading to $x_n=\frac{x_n-\overline{x_n}}{\max_d |x_n-\overline{x_n}|}$. 
The batch size is taken as $50$. 
Half of the batch is filled with labeled examples and the remaining with unlabeled ones draw randomly from $\mathcal{D}_u$.
One epoch corresponds to having treated all unsupervised examples. As such, supervised examples $\mathcal{D}_s$ is augmented by replicating the labeled examples as many times as necessary to obtain ${\rm Card}(\mathcal{D}_s)={\rm Card}(\mathcal{D}_u)$. This is standard technique for non GAN based semi-supervised settings\cite{patel2016probabilistic}.
Finally, the only hyper-parameter to cross-validate is the initial learning rate $\gamma^{(0)}$. For this, we tried the following learning rates $\gamma^{(0)} \in \{0.02,0.002,0.0002\}$ and chose the greatest one which did not lead to DNN divergence during learning. Hence we use  $\gamma^{(0)}=0.002$ for all models and all experiments, with adam optimizer \cite{kingma2014adam}.
When $e=150$ and $e=300$ we perform a manual learning rate change by setting $\gamma^{(0)} = \gamma^{(0)} /2$ with $e$ denoting the epoch number. We train for a total of $400$ epochs.
Because we consider as one layer the succession of Convolution-Nonlinearity-Pooling for the CNN topology and a full block for the Resnet, our formula can be applied directly from Eqs.~\ref{eq_renorm},\ref{eq_phis}.

By using the previously described training settings, we now present results on the two datasets SVHN and CIFAR10. For both we use benchmarks to compare our results with two regimes : $500$ and $1000$ labels for SVHN, and $4000$ and $8000$ labels for CIFAR10.
We also compare the two introduced renormalized loss we denote as $\Gamma$ and $\lambda$, the latter being the one per layer.

\begin{table}
\caption{SVHN dataset error comparisons for local $\lambda$ versus multiscale $\Gamma$ losses. 
We see that at $\mathbf{N_L}=500$ then ${\rm err(\lambda)} = {\rm err(\Gamma)} \ll {\rm err(SotA)}$, and at $\mathbf{N_L}=1000$ then ${\rm err(\lambda)} < {\rm err(SotA)} < {\rm Err(\Gamma)}$. We present for the best model average over $12$ runs ($4$ for the sigmoid case).}
\centering
\begin{tabular}{|c|c|c|} \hline
$\mathbf{N_L} $              & $\mathbf{500}$   & $\mathbf{1000}$ \\ \hline
Large CNN $\lambda$           &     $10.17$    &    $7.20$   \\ \hline
Deep Resnet $\lambda$          &    $11.08$   &     $8.28$  \\ \hline
Wide Resnet $\lambda$           &   $\textbf{9.82}\pm 1.5$     &    $\textbf{7.14}\pm0.3$  \\ \hline
Mean with $\lambda$ & $ 10.35 \pm 0.53$  & $ 7.54 \pm 0.52 $ \\ \hline \hline
Large CNN $\Gamma$           &     $11.67$  &    $9.51$   \\ \hline
Wide Resnet $\Gamma$          &  $12.36$     &   $10.17$\\ \hline 
Deep Resnet $\Gamma$          &      $8.95$ &     $11.27$ \\ \hline
Mean with $\Gamma$                        & $10.99 \pm 1.80   $           & $10.32 \pm 0.89  $ \\ \hline \hline
Sigmoid Wide Resnet $\lambda$&$20.35\pm 6.2$&$7.78\pm0.2$ \\ \hline \hline
Improved GAN \cite{salimans2016improved}        & $18.44 \pm 4.8$ &$8.11 \pm 1.3$  \\ \hline
Auxiliary Deep &&\\
Generative Model \cite{maaloe2016auxiliary} & - &$22.86$ \\ \hline
Skip Deep&&\\
Generative Model \cite{maaloe2016auxiliary} & - & $16.61 \pm 0.24$ \\ \hline
Virtual Adversarial \cite{miyato2015distributional} & - & $24.63$ \\ \hline
DGN \cite{kingma2014semi} & - & $36.02 \pm 0.1$\\ \hline
Triple GAN \cite{li2017triple} &-&$94.23\pm0.17$\\ \hline
Semi-Sup Requires a&&\\Bad GAN \cite{dai2017good} &-&$95.75\pm0.03$\\ \hline
$\Pi$Model\cite{laine2016temporal} &$92.95\pm 0.3$&$94.57\pm0.25$\\ \hline
VAT\cite{miyato2015distributional} &-&$75.37$\\ \hline
\end{tabular}\label{svhn}
\end{table}

\begin{table}
\caption{CIFAR10 Dataset performances summary and comparison for local $\lambda$ versus global $\Gamma$ losses. 
for $\mathbf{N_L}=8000$ we observe ${\rm err(sigmoid CNN \lambda)< err(SotA) < err(\Gamma)}$.
For the best model we provide mean and standard deviation over $8$ runs for all cases.}
\centering
\begin{tabular}{|c|c|c|c|} \hline
$\mathbf{N_L} $              & $\mathbf{4000}$ & $\mathbf{8000}$\\ \hline
CNN $\lambda$         &   $22.63\pm 0.44$     &    $17.92 \pm 0.3$   \\ \hline 
Wide Resnet $\lambda$           &   $23.71$     &  $19.63$    \\ \hline
Deep Resnet $\lambda$          &    $28.64$    &  $24.53$    \\ \hline
Mean with $\lambda$             & $24.99 \pm 2.61 $&$20.69 \pm 2.8 $ \\ \hline \hline
CNN $\Gamma$          &   $25.19$    &      $19,96$ \\ \hline
Deep Resnet $\Gamma$           &  $27.62$     &    $21.99$ \\ \hline
Wide Resnet $\Gamma$           &  $26.18$     &  $21.24$ \\ \hline
Mean with $\Gamma$            & $26.33 \pm 0.79 $& $21.06 \pm 1.03 $ \\ 
\hline \hline
Sigmoid CNN $\lambda$&$21.91\pm0.42$&$\textbf{16.45} \pm 0.23$ \\ \hline \hline
Improved GAN \cite{salimans2016improved}        & $\textbf{18.63} \pm 2.32$ &$17.72 \pm 1.82$ \\ \hline
LadderNetwork \cite{rasmus2015semi} & $20.40 \pm 0.47$& -\\ \hline
catGAN \cite{springenberg2015unsupervised} & $19.58\pm 0.46$& -\\ \hline 
DRMM&&\\+KL penalty \cite{patel2016probabilistic}& $23.24$ &-\\ \hline
Triple GAN \cite{li2017triple} &$83.01\pm0.36$&-\\ \hline
Semi-Sup Requires a &&\\Bad GAN \cite{dai2017good} &$85.59\pm0.30$&-\\ \hline
$\Pi$Model\cite{laine2016temporal} &$83.45\pm0.29$&-\\ \hline
\end{tabular}\label{cifar}
\end{table}

We also provide the evolution of the training cross-entropy and entropy losses as well as reconstruction loss per layer and the test set accuracy evolution highlighting the fast convergence of the models in Fig.~\ref{losscifar},\ref{losscifarsigmoid}.

We analyze the reconstruction of the best model for each dataset in Fig.~\ref{rec}. While the $\Gamma$ loss provides DNNs with much better reconstruction capacities, the use of this ability for semi-supervised classification task is nonexistent. On the opposite as seen in Fig.~\ref{losscifar}, the inner layer's ability to reconstruct reduces dramatically. Yet, those inner representations are the ones of interest encoding the crucial information about the input, filtered after the cascade of projections and nonlinearities.
Thus the difference observed by changing from a global loss to a multiscale one, brings greater representation learning of the DNN and this regardless of the number of labeled samples. In fact, the reconstruction being applied for labeled and unlabeled examples, only the total number of samples will impact the observed benefits in accuracy.

We also present complementary experiment in order to highlight the ability of the proposed method to generalize not only between DNN architectures but also when changing nonlinearities. 
We already demonstrated the ability of the technique to deal with dropout and batch normalization as well as different tasks with no change in the framework. For this experiment we use the best model on the CIFAR 10 and SVHN tasks being respectively the CNN and wide resnet. Then we simply replace all nonlinearities originally being leaky rectifiers by sigmoids. We then apply the exact same experiment as before with no change whatsoever. We perform the learning by $\lambda$ loss and compare with the original models for each dataset in Tab.~\ref{cifar},\ref{svhn} as well as provide for the CIFAR10 case evolution of the losses and test set accuracy in Fig.~\ref{losscifarsigmoid}.

Such experiments have some importance: First, being able to generalize to non piecewise affine activation function allows the use of such a framework for DNN architectures requiring squashing functions such as recurrent networks \cite{jozefowicz2015empirical,graves2013generating}, LSTM \cite{hochreiter1997long} and GRUs \cite{chung2014empirical}. Secondly, for more general task, one might consider to impose to specific behavior of the hidden layer representations such as saturation, upper bounded output and so on. Hence, nonconvex function might be of interest leading to the impossibility to use ReLU based activations.
For this specific experiment, one benefit of using a sigmoid function is the ability to bound the forward-backward pass. In fact, during backpropagation, the vanishing gradient property, usually considered as detrimental during learning, can here be considered as a self-regulatory behavior avoiding explosion of the reconstruction amplitudes through the layers.

\section{Discussion and Future Work}

In this paper, we have developed a general, parameterless loss function for learning. We demonstrated that it enables DNNs of any topology to be trained in a semi-supervised manner and that it is robust, leading to state-of-the-art performances on various tasks across DNN topologies and data sets.
By providing a general framework dealing with no task or DNN specific pre-processing; as well as being computationally efficient, we hope to bring DNNs to semi-supervised applications.

There are many avenues for future work.
For instance, we can see that the introduced multiscale loss ($\lambda$) is able to outperform the global loss ($\Gamma$). Yet, further improvements seem to be reachable with the standard trade off between computational need and model abilities. While we chose to present an out-of-the-box approach reaching state-of-the-art performances, there still remains options if one aims at further performances.
To do so, one possibility remains in the introduction of hyper parameters $(\beta_{CE},\beta_{E},(\beta_R^{(\ell)})_{\ell=0}^{L-1})$ in order to find more robust weighting of the losses such that the input reconstruction does not penalizes learning. However, doing so would bring back the cumbersome task of cross-validation. Hence one solution would be to do so coupled with automatic hyper-parameters updates as was done for learning rate.
To do so, two approaches would be available. First, as is done in adam \cite{zeiler2012adadelta,kingma2014adam} and rmsprop  \cite{tieleman2012lecture}, updates of the hyper-parameters based on their evolution through the updates, their statistics and behaviors could be used. Such as reducing the ones corresponding to a volatile loss or simply re-weighting the multiple losses to guarantees uniform speed of convergence among them. Secondly, a more explicit possibility would be to explicit optimize and update the hyper-parameters by line search or approximate line search with gradients as proposed in  \cite{jacobs1988increased,moreira1995neural}.
For example, this could take the form of
 updating the weighting while performing learning. Let index by $t$ the value of the parameters $\Theta$ at batch $t$. Given the new hyper-parameterized loss
\begin{align*}
    \mathcal{L}(x_n,y_n&;\Theta)=\beta_{CE}(t) \mathcal{L}_{CE}(y_n,\hat{y}(x_n))1_{\{y_n \not = \emptyset\}}\\
    &+\beta_E(t) \mathcal{L}_E(x_n)1_{\{y_n = \emptyset\}} +\sum_{\ell=0}^{L-1} \beta_R^{(\ell)}(t)\mathcal{L}_R^{(\ell)}(x_n),
\end{align*}
the updated weights are defined as
\begin{align*}
    \Theta(t+1)=\Theta(t) - \gamma g(\mathcal(x_n,y_n;\Theta),\Theta),
\end{align*}
with typically $g(\mathcal{L}(x_n,y_n;\Theta),\Theta)=\frac{d \mathcal{L}(x_n,y_n;\Theta)}{d\Theta}$ being a gradient descent update. One can thus adopt the following update strategy for the hyper-parameters as
\begin{align*}
    \beta^{(\ell)}(t+1) = \beta^{(\ell)}(t)-\frac{d \mathcal{L}(x_n,y_n;\Theta(t+1))}{d\beta^{(\ell)}(t)},
\end{align*}
and so for all hyper-parameters.
Finally, from another angle, studying the impact of batch size as was done for supervised learning \cite{patel2017impact,keskar2016large} as well as the proportion of labeled versus unlabeled examples per batch is of crucial important to further provide robust yet adaptive learning of large scale networks.


{\small
\bibliographystyle{ieee}
\bibliography{egbib}
}

\end{document}